# Object Localization Assistive System Based on CV and Vibrotactile Encoding

Zhikai Wei, Xuhui Hu

*Abstract—* Intelligent assistive systems can navigate blind people, but most of them could only give non-intuitive cues or inefficient guidance. Based on computer vision and vibrotactile encoding, this paper presents an interactive system that provides blind people with intuitive spatial cognition. Different from the traditional auditory feedback strategy based on speech cues, this paper firstly introduces a vibration-encoded feedback method that leverages the haptic neural pathway and enables the users to interact with objects other than manipulating an assistance device. Based on this strategy, a wearable visual module based on an RGB-D camera is adopted for 3D spatial object localization, which contributes to accurate perception and quick object localization in the real environment. The experimental results on target blind individuals indicate that vibrotactile feedback reduces the task completion time by over 25% compared with the mainstream voice prompt feedback scheme. The proposed object localization system provides a more intuitive spatial navigation and comfortable wearability for blindness assistance.

## I. Introduction

Vision provides highly accurate and detailed spatial information about the three-dimensional properties of external objects [1]. It was estimated that there were 43.3 million blind or visually impaired (BVI) individuals in 2020 [2]. They usually have difficulties in outdoor activities (e.g., walking in a complex and crowded place or avoiding obstacles and holes on the road) and many indoor tasks (e.g., picking up a cup from an unfamiliar desk or recognizing clothes with certain colors). Nowadays, a variety of assistance systems have been developed to navigate BVI individuals [3-11]. These systems mainly focus on two forms of feedback: audio feedback and physical force feedback. Audio feedback often uses sound effects or text-to-speech to inform the user of the location of obstacles or the destination [3-7]. Though this method is easy for users to understand, it has limitations in providing detailed information of complex environments during navigation, because a large amount of text of the speech and a high frequency of sound effects is required to describe the details of the navigation route. In [8-9], the phase difference between the left and right ear sounds was exploited to make the impression of directionality. The stereo headphones can provide distinct information of left and right directions, but they have difficulties in establishing a forward and backward sense of direction. Although these methods can navigate BVI individuals, they are still unable to describe detailed environmental information, and the users have to passively receive the navigation path planned by the robots. Moreover, the users need to manipulate the robot manually all the time, resulting in a lack of portability and wearability.

Haptic navigation systems can solve these problems. S. Ertan et al. [10] proposed a wearable haptic navigation guidance system using a 4-by-4 array of micromotors mounted in the back of a vest. J. H. Hogema et al. [11] designed a haptic rendering algorithm by utilizing neck perception. This paper aims to develop a wearable navigation system that can locate target objects and the user's hand and establish intuitive spatial perceptions for the users to catch the objects. To achieve this goal, a system based on computer vision and vibrotactile encoding is proposed. Object detection is achieved by using a head-mounted RGB-D camera. Meanwhile, to acquire the 3D spatial locations of the objects and the user's hand, a posture sensor is added to transform the camera coordinate system to the user's coordinate system. In terms of the feedback method, with the location information transmitted from the vision module, the direction and the distance of the user's hand to the target object are encoded into the vibration sequence of four motors. With the real-time vibrotactile feedback, the users can build the spatial cognition and easily find a reaching route to the target object. Encoded vibrotactile feedback is exploited because it takes advantage of the sensitive vibrotactile perception of BVI individuals. Also, with the collaboration of multiple vibration motors, the system can provide intuitive information of complex environments for the user. Moreover, it frees up the user's hands and improves the wearability of the device.

## II. Methods

### A. System Framework and Setup

Fig. 1(a) shows the framework of the developed blindness assistance system. It contains a vision module, a controller, and a vibrotactile feedback module. Firstly, the vision module acquires the external environment information. Then, the controller identifies and localizes the target object. Afterward, a sub-controller in the feedback module receives the location information and provides encoded vibrotactile information for the user.

Fig. 1(b) presents the system setup. The vision module is composed of a camera and a posture sensor. In this study, the RealSense D435 camera is used as the vision sensor, which provides RGB image and a point cloud data with wide field of view. Since the camera coordinate system will change with the rotation of the user's head, a posture sensor containing an

This work was supported by National Natural Science Foundation of China (No. 91648206), and Basic Research Program of Jiangsu Province (BK201900240).

Z. Wei, A. Song and X. Hu are with the State Key Laboratory of Bioelectronics and Jiangsu Key Laboratory of Remote Measurement and Control, School of Instrument Science & Engineering, Southeast University, Nanjing Jiangsu, 210096, China.

Correspondence to Email: a.g.song@seu.edu.cn; Tel.: +86 13951804055.

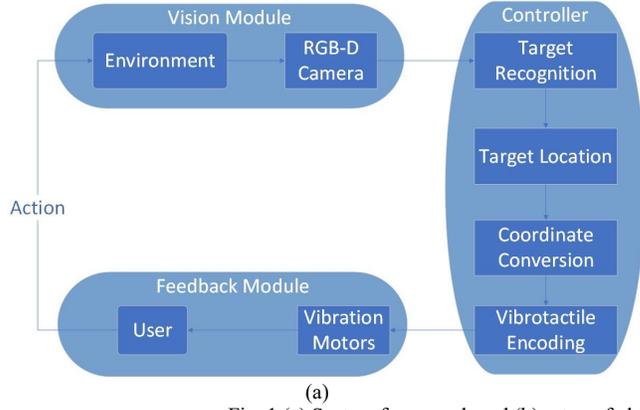 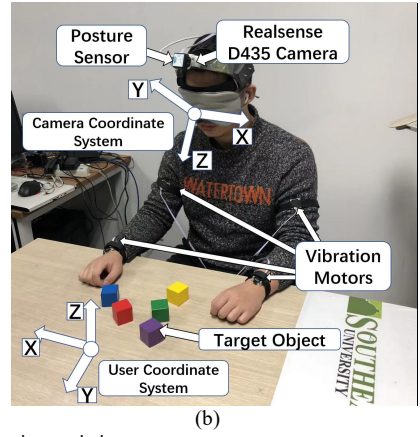

Fig. 1 (a) System framework and (b) setup of object localization assistive system inertial measurement unit is added to calibrate the object's orientation. Meanwhile, a minicomputer is placed on the user's waist as a controller of the system, and it links the vision module and the feedback module with a wired connection. Then, the feedback module acquires the relative orientation of the target to the user's hand, and an STM32F103C8T6 (ST Ltd.) micro-controller unit is placed on the user's waist as a sub-controller to output pulse width modulation (PWM) signals to the vibration motors on the user's arms.

*B. Object Detection Based on Computer Vision*

The vision module uses computer vision to identify and locate the user's hand and the target object with the assistance of three frameworks, namely "Object Detection", "Box Tracking", and "Hands" in MediaPipe (Google Inc.). The module supports cross-platform deployment and can be used on mobile devices. To ensure the portability of the developed device, a mini-computer (GMK NucBox) is used as the controller. The computer is a pocket-sized x86 platform equipped with an Intel J4125 CPU (2.7GHz, 4 cores with 4 threads.

Firstly, the camera obtains the pixel coordinates of the target object and the user's hand in the RGB image. Then, the controller aligns the RGB image with the depth image to obtain the 3D spatial coordinates in the camera coordinate system. Afterward, the posture sensor acquires the quaternion: $Q = q_0 + q_1 i + q_2 j + q_3 k$. Based on the rotational transformation matrix shown in (1), the point cloud coordinates in the camera coordinate system are converted to those in the user coordinate system, as shown in (2). The user coordinate system and the camera coordinate system are defined in Fig. 1(b). It can be seen that under the user coordinate system, the direction of each axis is consistent with the user's spatial perception.

$$C_b^n = \begin{bmatrix} q_0^2+q_1^2-q_2^2-q_3^2 & 2(q_1q_2-q_0q_3) & 2(q_1q_3+q_0q_2) \\ 2(q_1q_2+q_0q_3) & q_0^2-q_1^2+q_2^2-q_3^2 & 2(q_2q_3-q_0q_1) \\ 2(q_1q_3-q_0q_2) & 2(q_2q_3+q_0q_1) & q_0^2-q_1^2-q_2^2+q_3^2 \end{bmatrix} \quad (1)$$

$$x_{ucs} = C_b^n \cdot x_{ccs} \quad (2)$$

*C. The Vibration Feedback*

The design of the feedback module involves three aspects: including the type of vibration motors, the vibrotactile encoding method, and the layout of the motors.

The selection of the vibration motor's type significantly affects the wearing comfort and vibration perception. In this study, two types of motors are compared. The first one is the circular flat motor. Its temperature will increase rapidly after running for a while, and it has a small vibration amplitude. The other one is the rotor motor. Because of its high control accuracy and large vibration amplitude, the rotor motor is used in this study, and a 3D printed shell is designed to isolate it from the human skin, thus ensuring comfort and safety.

In terms of the vibrotactile encoding method, this study mainly focuses on the use of multiple motors to feedback directions and distance. The schematic of the vibrotactile encoding is shown in Fig. 2(a). The black arrow represents the direction vector from the user's hand to the target in the user's coordinate system, where the distance and the angle are denoted as $d$ and $\theta$, respectively. The four motors represent four directions (m1 for right, m2 for front, m3 for left, m4 for back). TABLE I lists the proposed encoding method, where $d_{max}$ represents the longest distance the system can detect and the duty cycle of PWM for each motor. Firstly, when the target is on one side of the hand, the corresponding motor starts to

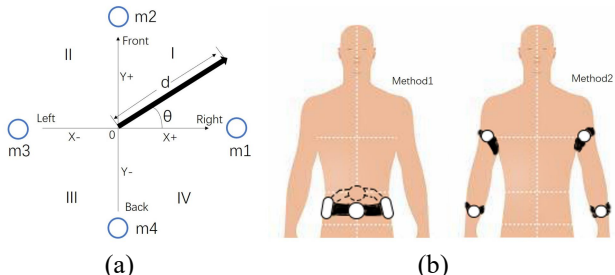

Fig. 2 (a) Schematic of the vibrotactile encoding. (b) Different layouts of vibration motors vibrate. Since the user may move the hand to the oblique side of the target, there will be two motors running at the same time. When the user's hand is moving towards the target, the closer the distance is, the higher the intensity of vibration will be. When the user's hand moves from the oblique side to the orthogonal side of the target, only the motor corresponding to the target's orientation keeps vibrating. This method provides a simple but obvious stimulation so that the user can easily understand the feedback information.

TABLE I. Vibrotactile Encoding Method

| PWM | X+ | I | Y+ | II | X- | III | Y- | IV |
|---|---|---|---|---|---|---|---|---|
| m1 | $\frac{(d_{max} - d \times \cos\theta)}{d_{max}} \times 100\%$ | | | 0 | | | | $\frac{(d_{max} - d \times \cos\theta)}{d_{max}} \times 100\%$ |
| m2 | 0 | $\frac{(d_{max} - d \times \sin\theta)}{d_{max}} \times 100\%$ | | | | 0 | | |
| m3 | | | 0 | | $\frac{(d_{max} + d \times \cos\theta)}{d_{max}} \times 100\%$ | | | 0 |
| m4 | | | | 0 | | | $\frac{(d_{max} + d \times \sin\theta)}{d_{max}} \times 100\%$ | |

As each part of the human body has different perception thresholds for vibration, the placement of vibration motors needs to ensure obvious vibration perception. In our preliminary experiment, an intuitive method was investigated (Fig. 2(b) Method 1). However, since the human abdomen has a high threshold for vibration perception, the vibration motor did not have a good effect. To address this issue, this study fixes two motors separately on each side of the upper arm to represent the left and right directions. Meanwhile, the other two motors are put on the two wrists (Fig. 2(b) Method 2). In this way, the users are provided with intuitive vibration perception.

## III. EXPERIMENTS

In this study, experiments were conducted to compare the vibration feedback method (VB) with the current popular voice prompt feedback method (VP). The experiments were divided into two sessions, i.e., a virtual interaction evaluation experiment and a real scene experiment.

Four sighted people (25.5 ± 4.2 years old) and three BVI individuals (22.6 ± 1.5 years old) participated in the experiment, and all experiments were approved by the University Ethics Committee and in accordance with the Declaration of Helsinki. The sighted subjects were blindfolded through the entire experiment, so they cannot observe and perceive the environment with their hands in advance.

To avoid the influence of the vision module, this study designed a virtual interaction experiment to focus on the feedback method. VP informed the subjects of orientation and distance information by guiding words: "forward", "backward", "left", and "right". The guiding words were played at an interval of six seconds to prevent the users from hearing fatigue as well as alert of a certain frequency, including 0.4Hz, 1Hz, and 2Hz. As for the VB method, the guidance strategy was designed as two stages, and the users were respectively guided to align horizontally and vertically to the target in the two stages. Meanwhile, a virtual interaction software based on Unity3D was designed for the experiment. The experiment required the subject to move the mouse cursor to reach a red circular endpoint on a two-dimensional plane in the virtual scene. When the cursor reached the endpoint, the trial was completed. At the beginning of each trial, the endpoint appeared at a random position.

The setup of the real-scene experiment is shown in Fig. 1(b). The subject wore a visual perception module on his head and sit directly in front of a table. Five blocks of different colors with the same mass and volume were placed on the table so that the subject cannot distinguish the objects only by touching them. At the beginning of each trial, the tester randomly disrupted the blocks and let them maintain a distance of 5-6 cm from each other. There were totally 340

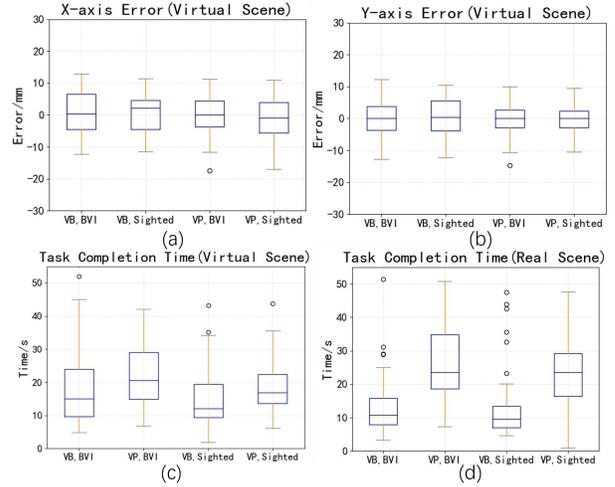

Fig. 3 (a) X-axis error (virtual scene), (b) Y-axis error (virtual scene), (c) Task completion time (virtual scene), (d) Task completion time (Real scene) trials in both experiments (3 subjects × 30 trials ×2 sessions for BVI subjects and 4 subjects × 20 trials × 2 sessions for sighted subjects). In the experiments, the success rate, task completion time, and navigation accuracy were evaluated, and the reaching path of each trial was recorded to evaluate the goodness of the navigation route.

TABLE II. Task Completion Time

| Time/s | | Virtual Experiment | | Real Scene Experiment | |
|---|---|---|---|---|---|
| | | Mean | Standard deviation | Mean | Standard deviation |
| VB | Sighted Subjects | 15.8 | 11.1 | 11.8 | 8.3 |
| | BVI Subjects | 20.9 | 17.5 | 13.0 | 7.7 |
| VP | Sighted Subjects | 18.3 | 7.6 | 25.1 | 12.9 |
| | BVI Subjects | 24.2 | 16.5 | 26.1 | 11.5 |

## IV. RESULTS

### A. Virtual Experiment

The data was collected from 340 trials, and the success rate of both feedback methods exceeded 90% (VB: (95.5 ± 0.18)%, VP: (98.7 ± 0.08)%). In this study, the positioning error of this experiment was defined as the distance between the cursor and the circle center of the endpoint at the end of each completed trial. The positioning error in the x-axis direction and the y-axis direction is illustrated in Fig. 3(a) and Fig. 3(b), respectively. In terms of efficiency, the task completion time of the two feedback methods is shown in Fig. 3(c) and TABLE II. The white circles represent the outliers ($Q_o > Q_u + 1.5 \times IQR$ or $Q_o < Q_L - 1.5 \times IQR$, $Q_o$: Outlier, $Q_u$: Upper quartile, $Q_L$: Lower quartile, $IQR$ : InterQuartile Range). It can be seen that the BVI subjects spent similar time with the sighted subjects, and the task completion time of VB was reduced by over 15% compared to VP for both sighted subjects and BVI subjects.

### B. Object Manipulation Experiment of Real Scene

The data of the 340 trials in the real-scene experiments were analyzed. The success rate of VB and VP was similar

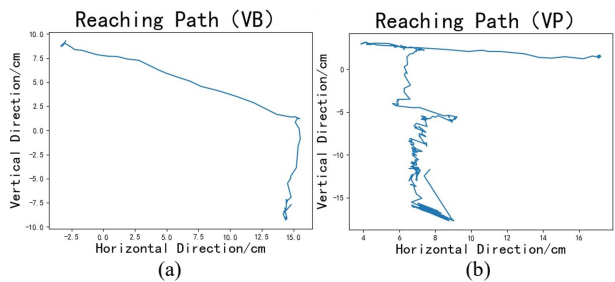

Fig. 4 (a)Reaching path of VB, (b)Reaching path of VP

(VB: (81.2 ± 0.4)%, VP: (80.8 ± 0.39)%). The main differences were in the reaching path and the task completion time. As shown in Fig. 4(a), the representative reaching path of VB is quite natural, while that of VP (Fig. 4(b)) contains many right-angle movements on both horizontal and vertical directions because the subject's moving was unconsciously skewed and corrected several times by voice prompt. Compared with VP, VB assists the subject to plan a more natural and smoother route, which saves time and provides users with stronger and more intuitive spatial perception and clearer orientation cues. Fig. 3(d) and TABLE II show that the mean task completion time of VB ranges from 11.8s (sighted) to 13.0s (BVI) and that of VP ranges from 25.1s (sighted) to 26.1s (BVI). This indicates that VB can cut down the task completion time by over 50% compared to VP. The outliers much disperse for VB because subjects were not familiar with the feedback at first ant it took several times of trials for them to get used to it.

## V. Discussion and Conclusion

The experimental results have shown that there is little difference between the success rate of VB and VP, and the reason for this is two-fold. One reason is that both methods provide subjects with complete orientation information; the other reason is that there was no strict time limit during the experiment, so the subjects only needed to finish the task no matter how long it took.

Efficiency is a more important evaluation metric. Both the virtual interaction experiment and the real-scene experiment present the same result that VB achieves higher positioning accuracy than VP. Meanwhile, VB could improve the task completion time by 15% and 50% in the virtual interaction experiment and the real-scene experiment, respectively. Thus, VB is significantly more efficient.

The observation of the experiment processes and the test data indicates that there are several reasons for the longer task completion time of VP. Firstly, VP plays the guiding words within a certain time interval, so the subjects need to wait a while for the guiding words and then make the next movement. Secondly, when the object is exactly in the orthogonal direction of the subject's hand, the subjects only need to move his hand unidirectionally, and such a route is similar to the optimal path. However, if the target is on the oblique side, the subjects could only move rigidly along the horizontal direction first and then along the vertical direction under the VP's guidance. Lastly, compared with VP, VB provides the subjects with intuitive feedback rather than a rigid navigation path, so the subjects can make decisions proactively and achieve a trajectory closer to the optimal navigation route.

Since this study only uses four motors currently, there are still some limitations. Voice interaction function will be added for users to operate the system. The vibrotactile encoding method will be improved in the future study by using more motors and more sophisticated control algorithms to offer 360-degree wrap-around uniform vibration feedback.

In conclusion, this paper presents an objective localization assistive system based on computer vision and vibrotactile encoding to provide intuitive environment perception for BVI individuals. The computer vision based on deep learning is exploited to recognize objects, and a depth camera is taken to collect 3D spatial information. Besides, a minicomputer is adopted as the controller, which has great portability and wearability. This method takes full advantage of the human body's high sensitivity to tactility and can express intuitive environmental information. It achieves higher positioning accuracy and efficiency than the mainstream voice prompt feedback method.